\documentclass[letterpaper, 10 pt, conference]{ieeeconf}
\IEEEoverridecommandlockouts                              

\usepackage{times}
\usepackage{multicol}
\usepackage[bookmarks=true]{hyperref}
\usepackage{graphics} 
\usepackage{amsmath} 
\usepackage{amssymb}  
\usepackage{mathtools}
\usepackage{float}
\usepackage{algpseudocode, algorithmicx, algorithm}
\usepackage[font=small,labelfont=bf]{caption}
\usepackage{graphicx}
\usepackage{multirow}
\usepackage{bm}
\usepackage{booktabs}
\usepackage{array}
\usepackage{threeparttable}
\usepackage[dvipsnames]{xcolor}
\usepackage{cite}
\usepackage{subfig}
\usepackage[stable]{footmisc}

\hypersetup{
    colorlinks=true,
    linkcolor=blue,
    filecolor=magenta,      
    urlcolor=cyan,
    pdfpagemode=FullScreen,
}

\DeclareMathAlphabet{\mathsfit}{T1}{\sfdefault}{\mddefault}{\sldefault}
\SetMathAlphabet{\mathsfit}{bold}{T1}{\sfdefault}{\bfdefault}{\sldefault}

\newcommand{\qcmd}[1][]{q_{\text{cmd}}^{#1}}
\newcommand{\qref}[1][]{q_{\text{ref}}^{#1}}
\newcommand{\tauE}{\tau_{\text{ext}}}
\newcommand{\tauM}{\tau_{\text{m}}}
\newcommand{\norm}[1]{\left\lVert{#1}\right\rVert}

\newcommand{\quantity}[4]{{}^{#2}{#1}^{#3}_{#4}} 

\renewcommand{\baselinestretch}{0.975}

\title{\LARGE \bf
Easing Reliance on Collision-free Planning with Contact-aware Control}

\author{Tao Pang and Russ Tedrake \\ 
\texttt{\{pangtao, russt\}@csail.mit.edu}
\thanks {This work is supported by Navy-ONR Award N00014-18-1-2210 and Lincoln Laboratory Award PO\# 7000470769.}%
}

\begin{document}
\maketitle
\thispagestyle{empty}
\pagestyle{empty}

\begin{abstract}
We believe that the future of robot motion planning will look very different than how it looks today: instead of complex collision avoidance trajectories with a brittle dependence on sensing and estimation of the environment, motion plans should consist of smooth, simple trajectories and be executed by robots that are not afraid of making contact.
Here we present a ``contact-aware" controller which continues to execute a given trajectory despite unexpected collisions while keeping the contact force stable and small. 
We introduce a quadratic programming (QP) formulation, which minimizes a trajectory-tracking error subject to quasistatic dynamics and contact-force constraints.
Compared with the classical null-space projection technique, the inequality constraint on contact forces in the proposed QP controller allows for more gentle release when the robot comes out of contact.
In the quasistatic dynamics model, control actions consist only of commanded joint positions, allowing the QP controller to run on stiffness-controlled robots which do not have a straightforward torque-control interface nor accurate dynamic models.
The effectiveness of the proposed QP controller is demonstrated on a KUKA iiwa arm. 
Project video: \url{https://youtu.be/M-7JMQRkiPk}.
\end{abstract}

\section{Introduction}
\begin{figure}
    \centering
    \includegraphics[width=0.65\linewidth]{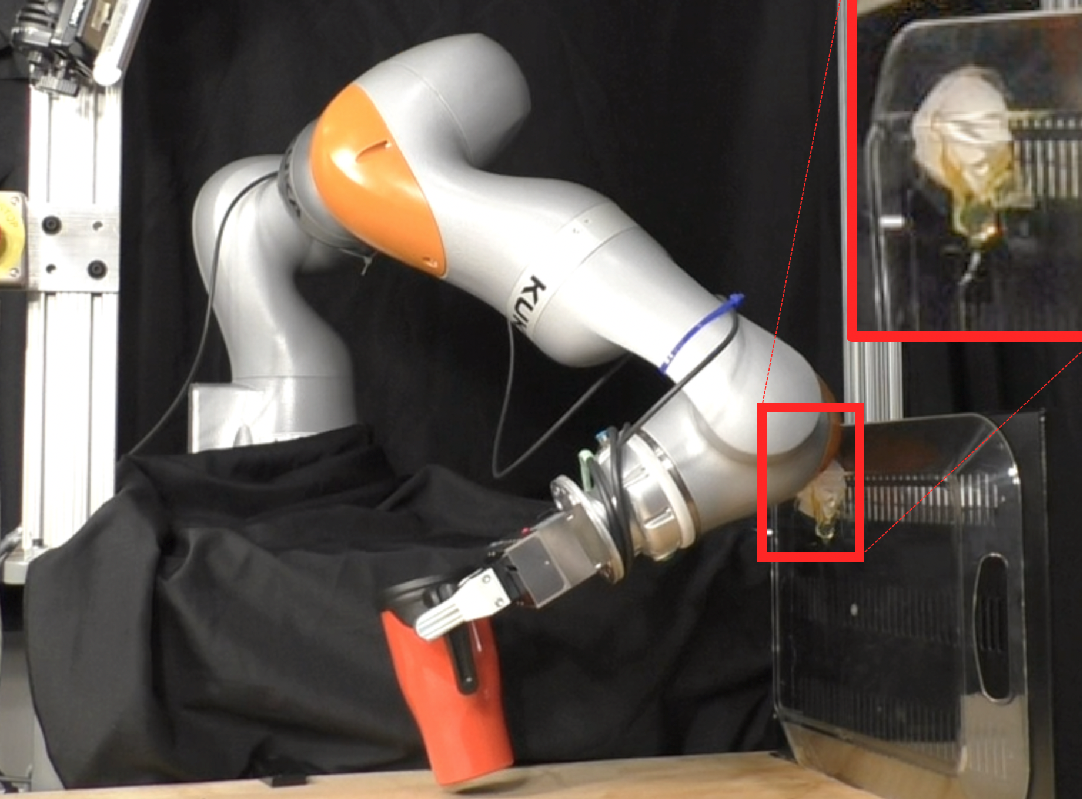}
    \caption{As transparent obstacles are almost invisible to depth sensors, a collision-free motion planner, with the goal to pick up the mug and unaware of the transparent tray, plans a trajectory that crushes the egg along the way. The crushed egg is highlighted in the red box. Our controller is able to keep the egg intact even when the reference trajectory would crush it.}
    \label{fig:contact_crushes_egg}
    \vspace{-0.7cm}
\end{figure}

Most robots today are programmed to move through the world as if they are afraid of making contact. Perhaps they should be: unexpected collisions while a robot is tracking a trajectory can create a large force at the point of contact, putting at risk both the robot and the environment with which it interacts (Fig. \ref{fig:contact_crushes_egg}). Consequently, a significant amount of effort and care in robot motion planning is spent on avoiding collisions. For example, sampling-based planners need to perform numerous collision checks \cite{lavalle2006planning}; optimization-based planners need to constantly evaluate the signed distance functions and their gradients \cite{ratliff2009chomp}. Moreover, high resolution collision geometries are usually needed to increase the chances of finding a collision-free path, which further increases the computational cost \cite{pan2012fcl}.

However, the complete and total avoidance of contacts is a severe limitation, even if we are willing to tolerate the computational cost. First, the effectiveness of collision-free planning is limited by the quality of the geometric models used for collision checks. Unless in structured environments where everything has been perfectly measured, models of the environment need to be reconstructed from range sensor (e.g. depth camera) measurements, which usually have a fair amount of uncertainty and can suffer from occlusion. Moreover, collision-free trajectories can be unnecessarily conservative \cite{mason2018toward}: a task achievable by making some contacts can be deemed infeasible by a collision-free planner. 

In this work, we propose a QP controller which, given estimated contact positions and forces in unexpected contacts, tracks the reference trajectory as closely as possible while keeping contact forces below a user-defined upper bound.
Compared with similar controllers based on null-space projection \cite{nakamura1987task, siciliano1991general, aghili2005unified, dehio2018modeling, jorda2019contact}, the QP formulation shares the same underlying dynamics, but allows for more gentle separation when the robot breaks contact with the environment. The gentle separation happens naturally as a result of bounding the contact forces with inequality constraints, which are not supported by null-space projection.

Instead of the usual second-order dynamics constraints used in robot locomotion \cite{kuindersma2014efficiently, koolen2016design}, the proposed QP controller utilizes a quasistatic dynamics model which predicts future equilibrium configurations and contact forces for a stiffness-controlled robot in response to position commands. By assuming bi-lateral, frictionless contacts, the quasistatic dynamics model can be expressed as equality constraints and does not require the estimation of friction coefficients. Moreover, as real-world contacts are uni-lateral and frictional, we also propose measures which both capitalize the simplicity of bi-lateral, frictionless contact models and mitigate the side effects of modeling real-world contacts as such.

\section{Related Work  \label{sec:related_work}}
\subsection{Interaction control}
As the primary objective of the proposed QP controller is to bound unexpected contact forces while tracking a joint-space or end-effector trajectory, we review existing methods for combined motion and force control, which are also referred to as interaction control in the literature \cite[Chapter~9]{siciliano2008springer}. Interaction control techniques can be broadly classified by whether the interaction force is controlled \textit{directly} or \textit{indirectly}. 

Direct force control typically splits the task space into two orthogonal subspaces based on the robot's kinematic constraints: one motion-controlled subspace along the tangents of the kinematic constraints, and one force-controlled subspace along the normals. Desired motion and force trajectories are specified in the motion and force controlled subspaces, and tracked independently using motor torque commands computed from the robot's second-order model \cite{mason1981compliance, khatib1987unified, raibert1981hybrid}. 

However, the success of direct force control relies on accurate robot and environment models, which are not easily available in unstructured environments. Moreover, with few exceptions \cite{franka}, most industrial robot arms, including the KUKA iiwa, do not have an interface for end-users to directly control motor torques \cite{kuka2019fri}. Last but not least, by directly controlling motor torque, the interaction controller bypasses the robot's factory motion controller, and thus needs to run at high frequency in order to maintain stability.

On the other hand, a classical example of indirect force control is impedance control \cite{hogan1984impedance}, which regulates the robot's response to external forces to that of a mass-spring-damper system (a mechanical impedance), thereby guaranteeing interaction stability by passivity. When the robot moves slowly, which is often the case in manipulation tasks, impedance control can be simplified to stiffness control \cite{salisbury1980active}, which can be interpreted as connecting the robot's end effector to a user-specified set-point by virtual springs. In the presence of contact, contact force can be controlled by commanding how much the set point penetrates the obstacle.

Compared with direct force control, indirect force control schemes are usually implemented as an outer-loop around the robot's factory motion controller, and therefore does not bear the responsibility of maintaining stability and can run at a much lower rate.

\subsection{Quasistatic dynamics models in manipulation}
Quasistatic dynamics has been used with great success to simplify the planning and control of simple tasks such as planar pushing \cite{hogan2020feedback}, where modeling robots as prescribed motion trajectories is sufficient. However, for multi-contact tasks such as grasping \cite{pang1996complementarity}, such simplifications can lead to non-unique contact forces  or violation of non-penetration constraints \cite{halm2018quasi, pang2018robust}. By modeling robots as impedances, our recent multibody quasistatic model \cite{pang2021convex} can faithfully reproduce the steady-state behavior of stiffness-controlled robot arms in multi-contact scenarios.

The quasistatic robot dynamics model proposed in this work, which predicts future joint angles in whole-arm contact scenarios, is an extension to classical indirect force control schemes that typically focus solely on the relationship between end effector pose and wrench \cite{salisbury1980active, hogan1984impedance}. The proposed quasistatic dynamics is also a frictionless simplification of the Coulomb-friction-based quasistatic dynamics \cite{pang2021convex}, which can be implemented on hardware with minimal contact sensing.

\subsection{Null-space projection}
Null-space projection is a classical and popular technique for executing a hierarchy of tasks defined by equality constraints \cite{nakamura1987task, siciliano1991general, aghili2005unified, dehio2018modeling, jorda2019contact}. The constraints imposed by higher-priority tasks are enforced by projecting the torque needed by low-priority tasks into the null space of the higher-priority tasks. The projections are defined over the ranges and null spaces of the task Jacobians and their \textit{weighted} pseudo-inverses. It is noteworthy that the projections are generally not orthogonal, unless the weight matrix is identity \cite{dietrich2015overview}.

More recently, controllers based on constrained optimizations such as quadratic programs (QP) have gained popularity in both locomotion \cite{koolen2016design, kuindersma2014efficiently} and manipulation \cite{jain2013manipulation}. Compared with null-space projections, QP-based controllers can handle both equality and inequality constraints. In this work, we formulate the problem of trajectory tracking with bounded contact forces as a QP with a novel quasistatic dynamics constraint. We also show that a controller based on null-space projection implicitly enforces the same quasistatic dynamics constraint when the projection is \textit{stiffness-consistent} \cite{dietrich2015overview}.

\section{Background and Notations}
\subsection{Constrained Inverse Dynamics Control}
A popular controller in locomotion and manipulation is based on the following optimization-based formulation \cite{kuindersma2014efficiently, koolen2016design, wang2019impact}:
\begin{subequations}
\small
\label{eq:generic_mpc}
\begin{align}
&\underset{x^{l+1}, u^l}{\text{min.}} \; c_x(x^{l+1}) + c_u(u^l), \; \text{s.t.}\\
&x^{l+1} = f(x^{l}, u^l), \label{eq:generic_mpc:dynamics_constraint}
\end{align}
\end{subequations}
where $x^{l}$ and $u^{l}$ are the state and input at the current time step $l$, and $x^{l+1}$ is the state at the next time step, $l+1$. The controller (\ref{eq:generic_mpc}) picks an action $u^l$ that minimizes the (usually LQR-style) state and action cost, $c_x(\cdot)$ and $c_u(\cdot)$, subject to the dynamics constraint (\ref{eq:generic_mpc:dynamics_constraint}).

The most common choice for the dynamics constraint (\ref{eq:generic_mpc:dynamics_constraint}) is the Newton's Second Law (N2L). For example, the DLR (German Aerospace Center) family of robots, including the KUKA iiwa and FRANKA panda, has the following closed-loop second-order dynamics after gravity compensation\cite{ott2008passivity}: 
\begin{equation}
\small
\label{eq:iiwa_second_order_closed_loop_dynamics}
M(q) \ddot{q} + \left(C(q, \dot{q}) + D_q\right) \dot{q} + K_q(q - \qcmd) = \tauE,
\end{equation}
where $q \in \mathbb{R}^{n_q}$ is the joint angles of the robot, $C(q, \dot{q}) \dot{q}$ is the Coriolis force, $D_q$ is a diagonal damping matrix, $K_q$ is a diagonal stiffness matrix, $\qcmd$ is the commanded joint angles, and $\tauE$ is the torque by external contacts.

\subsection{Contact and multibody notations}
We consider rigid, point contacts in this work. The number of contacts the robot makes with the environment is denoted by $n_c$. Each contact point is denoted by $C_i$. The coordinates of the contact point relative to world frame, expressed in world frame is written as $\quantity{p}{W}{C_i}{} \in \mathbb{R}^3$; contact force at $C_i$ expressed in world frame is represented by $\quantity{f}{}{C_i}{W} \in \mathbb{R}^3$. 

Position Jacobian of the contact point $C_i$ relative to frame $W$, expressed in frame $W$, is denoted by $J_{q}^{\quantity{p}{W}{C_i}{}}(q) : \mathbb{R}^{n_q} \rightarrow \mathbb{R}^{3 \times n_q}$. It maps the robot's joint velocity $\Dot{q}$ to the velocity of point $C_i$ in world frame $W$:
\begin{equation}
\small
\quantity{v}{W}{C_i}{} = J_{q}^{\quantity{p}{W}{C_i}{}}(q) \Dot{q}.
\end{equation}

We further define
\begin{subequations}
\small
\label{eq:contact_definitions}
\begin{align}
f_i  &\coloneqq \norm{\quantity{f}{}{C_i}{W}} \in \mathbb{R};  \;  u_i \coloneqq \quantity{f}{}{C_i}{W} / f_i \in \mathbb{R}^3\\
J_{u_i} &\coloneqq u_i^{\intercal} J_{q}^{\quantity{p}{W}{C_i}{}} \in \mathbb{R}^{1 \times n_q}\\
J_u &\coloneqq \left[J_{u_1}^\intercal, J_{u_2}^\intercal, \cdots J_{u_{n_c}}^\intercal \right]^\intercal \in \mathbb{R}^{n_c \times n_q}. \label{eq:contact_jacobian}
\end{align}
\end{subequations}
The $i$-th row of $J_u$ maps $\Dot{q}$ to the Cartesian velocity of $C_i$ along $\bm{u}_i$ in world frame. We assume that $J_u$ is full-rank.

\section{Quasistatic Dynamics}
\subsection{Dynamics as transitions between equilibria \label{sec:quasistatic_dynamics_1}}
For robot arms with a joint-level stiffness controller, their steady-state equilibrium condition can be obtained by setting the derivative terms in the second-order dynamics (\ref{eq:iiwa_second_order_closed_loop_dynamics}) to 0:
\begin{equation}
\small
\label{eq:quaistatic_force_balance}
K_q (\qcmd - q) + \tauE = 0. 
\end{equation}

\begin{figure}[h]
\vspace{-0.3cm}
\centering
\includegraphics[width=0.45\linewidth]{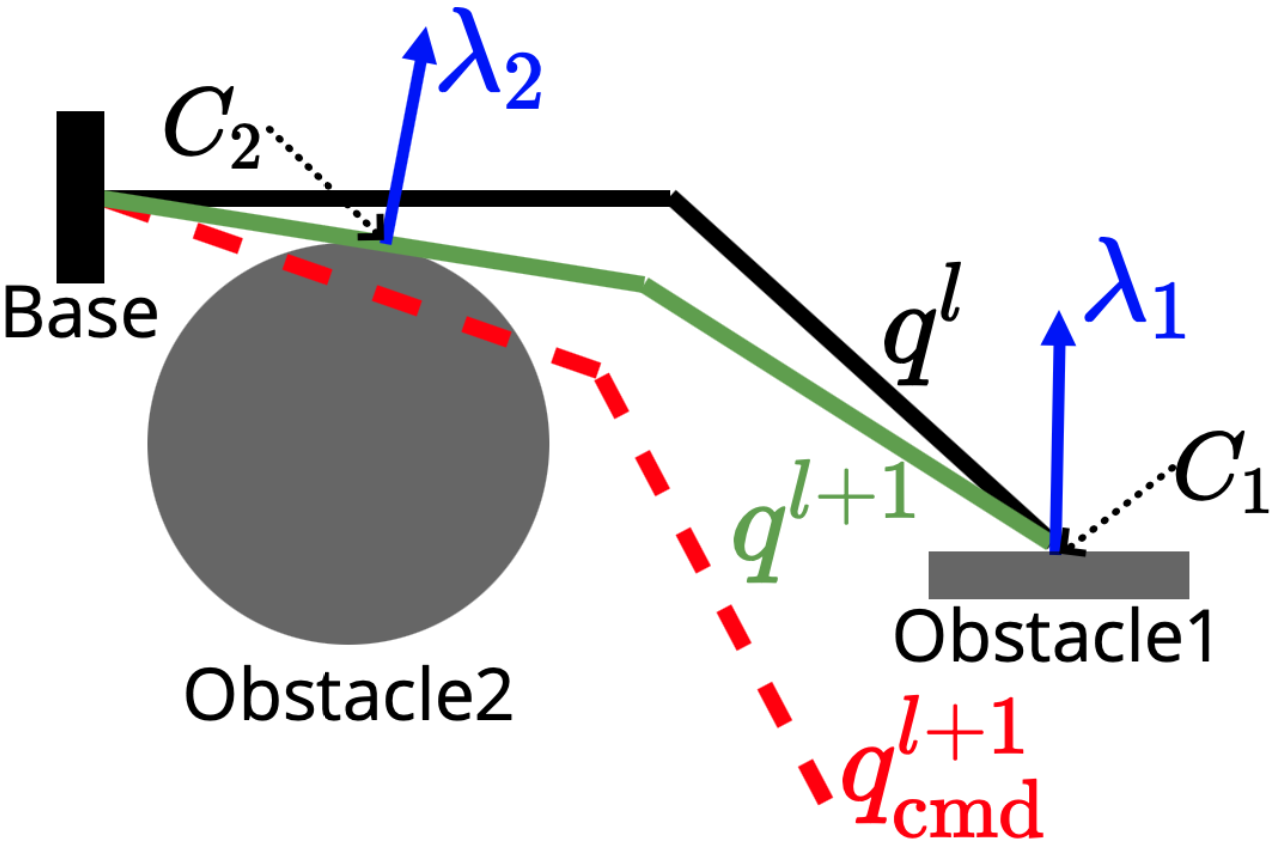}
\caption{Quasistatic dynamics of a 2D, 2-link robot arm. The arm starts at $q^l$ (black) and is commanded to go to $\qcmd[l+1]$ (red). The virtual spring connecting $\qcmd[l+1]$ to $q^l$ pulls the robot towards $\qcmd[l+1]$, but the robot eventually stabilizes to $q^{l+1}$ (green) due to contact constraints. At $l+1$, the arm makes two contact with two obstacles at $C_1$ and $C_2$ with contact forces $\lambda_1$ and $\lambda_2$.}
\label{fig:quasistatic_dynamics}
\vspace{-0.3cm}
\end{figure}

For a stiffness-controlled robot, we can define its \textit{quasi-static} dynamics, whose state consists only of the joint angles $q$, and input the commanded joint angles $\qcmd$. As shown in Fig. \ref{fig:quasistatic_dynamics}, the quasistatic dynamics predicts $x^{l+1} \coloneqq q^{l+1}$, the equilibrium configuration at the next time step, from the current equilibrium configuration $q^{l}$ and the next commanded configuration $u^l \coloneqq \qcmd[l+1]$.

The new equilibrium $q^{l+1}$ can be solved for by minimizing the potential energy of the robot, subject to the contact constraints:
\begin{subequations}
\small
\label{eq:quasistatic_dynamics_qp}
\begin{align}
&\underset{q^{l+1}}{\text{min.}} \; \frac{1}{2}(\qcmd[l+1] - q^{l+1})^\intercal K_q (\qcmd[l+1] - q^{l+1}) \; \text{s.t.} \\
& J_u(q^l) (q^{l+1} - q^{l}) = 0. \label{eq:quasistatic_dynamics_qp:equality_constraint}
\end{align}
\end{subequations}

To derive the contact forces, we start with the Lagrangian of QP (\ref{eq:quasistatic_dynamics_qp}):
\begin{equation}
\small
\begin{aligned}
L(q^{l+1}, \lambda) = &\frac{1}{2} (\qcmd[l+1] - q^{l+1})^\intercal K_q (\qcmd[l+1] - q^{l+1}) \\
&- (\lambda^{l+1})^\intercal J_u(q^{l+1} - q^l)
\end{aligned}
\end{equation}
where $\lambda^{l+1} \in \mathbb{R}^{n_c}$ is the Lagrange multipliers of the contact constraint (\ref{eq:quasistatic_dynamics_qp:equality_constraint}), which can also be interpreted as the contact forces generated by (\ref{eq:quasistatic_dynamics_qp:equality_constraint}); the dependency of $J_u$ on $q^l$ is dropped for simplicity.

The KKT optimality condition of QP (\ref{eq:quasistatic_dynamics_qp}) is given by
\begin{subequations}
\small
\label{eq:dynamics_qp_kkt}
\begin{align}
\nabla_{q^{l+1}}L = K_q (q^{l+1} - \qcmd[l+1]) - J_u^\intercal \lambda^{l+1} &= 0, \label{eq:dynamics_qp_kkt:gradient} \\
  J_u \left(q^{l+1} - q^l\right) &= 0. \label{eq:dynamics_qp_kkt:equality_constraint}
\end{align}
\end{subequations}
where (\ref{eq:dynamics_qp_kkt:gradient}) is equivalent to the steady-state force balance condition (\ref{eq:quaistatic_force_balance}), assuming that $\tauE$ is generated by the $n_c$ point contacts, i.e. $\tauE = J_u^\intercal \lambda^{l+1}$.

Explicit expressions for $\lambda^{l+1}$ and $q^{l+1}$ can also be derived from the KKT conditions (\ref{eq:dynamics_qp_kkt}):
\begin{subequations}
\small
\label{eq:explicit_force_and_state}
\begin{align}
\lambda^{l+1} &= -(J_u K_q^{-1} J_u^\intercal)^{-1} J_u (\qcmd[l+1] - q^l), \label{eq:explicit_force_and_state:force}\\
q^{l+1} &= q^l + \left(I -  K_q^{-1}J_u^\intercal (J_u K_q^{-1} J_u^\intercal)^{-1}J_u \right) (\qcmd[l+1] - q^l). \label{eq:explicit_force_and_state:state}
\end{align}
\end{subequations}

\subsection{Relationship to null-space projection}
In this sub-section, we show that when controlling stiffness-controlled robots using null-space projection, lower-priority tasks can be guaranteed to not interfere with higher-priority tasks during transients if the stiffness-consistent projection \cite{dietrich2015overview} is used. We also show that the underlying dynamics model of a controller based on stiffness-consistent projection is the same as the model proposed in Sec. \ref{sec:quasistatic_dynamics_1}. 

Null-space projection technique revolves around two projections: 
\begin{subequations}
\small
\begin{align}
P_R(W) &\coloneqq J_u^\intercal \left(J_u^{W +}\right)^\intercal, \\
P_N(W) &\coloneqq I - J_u^\intercal \left(J_u^{W +}\right)^\intercal,
\end{align}
\end{subequations}
where $J_u^{W+} \coloneqq W^{-1} J_u^\intercal (J_u W^{-1} J_u^\intercal)^{-1}$ is the pseudo-inverse weighted by a positive-definite $W$. The range and null space of $P_R(W)$ are respectively $R(J_u^\intercal)$ and $N\left(\left(J_u^{W+}\right)^\intercal\right)$, whereas the range and null space of $P_N(W)$ are reversed. Note that such projections can be defined for arbitrary task Jacobians, but we specialize to the contact Jacobian defined in (\ref{eq:contact_jacobian}) without loss of generality.

For any choice of $W$ and any joint torque $\tauM$, $P_R(W) \tauM \in R(J_u^\intercal)$ generates contact forces, whereas $P_N(W) \tauM$ generates no joint torque in $R(J_u^\intercal)$ after static equilibrium is reached. Choosing an appropriate $W$, however, can provide additional guarantees during the transient into this steady state \cite{dietrich2015overview}. For instance, the \textit{dynamically-consistent} pseudo-inverse $J_u^{M+}$\cite{featherstone1997load}, which uses the robot's mass matrix $M$ for $W$, ensures that 
\begin{equation}
\small
\label{eq:dynamic_consistency}
0 \equiv J_u M^{-1} P_N(M) \tauM, \; \forall \tauM.
\end{equation}
Assuming that the dominant effect of $P_N(M) \tauM$ during the transient is to generate acceleration, property (\ref{eq:dynamic_consistency}) guarantees that and the generated acceleration lies inside $N(J_u)$.

To determine the appropriate choice of $W$ when the dominant effect of the $P_N(W) \tauM$ during transient is to stretch/contract the virtual spring of a stiffness-controlled robot, we start at the instant $l^+$, immediately after sending the joint angle command $u^l = \qcmd[l+1]$ at time step $l$. At $l^+$, the joint torque can be expressed as 
\begin{equation}
\small
\label{eq:tau_l_stiffness}
\tauM^{l^+} = K_q(\qcmd[l+1] - q^l),
\end{equation}
which can be decomposed as 
\begin{equation}
\small
\label{eq:tau_l_decomposition}
\tauM^{l^+} = \underbrace{P_R(W) \tauM^{l^+}}_{\tau_R} + \underbrace{P_N(W) \tauM^{l^+}}_{\tau_N},
\end{equation}
where $\tau_R \in R(J_u^\intercal)$ generates contact forces, and $\tau_N \in N\left(\left(J_u^{W+}\right)^\intercal\right)$ generates a motion that needs to be in $N(J_u)$.

Combining (\ref{eq:tau_l_stiffness}) and (\ref{eq:tau_l_decomposition}) yields
\begin{equation}
\small
\label{eq:tau_l_combined}
K_q(\qcmd[l+1] - q^l) = \tau_R +\tau_N.
\end{equation}

At time instant $(l+1)^-$, when the equilibrium at time step $l+1$ is reached but $u^{l+1}$ has not been commanded, $\tau_N$ has generated a displacement and been dissipated by damping, but $\tau_R$, the generalized force due to contact, remains:
\begin{equation}
\small
\label{eq:tau_l+1_decomposition}
\tauM^{(l+1)^-} = \tau_R.
\end{equation}
Moreover, static equilibrium (\ref{eq:quaistatic_force_balance}) at $(l+1)^-$ dictates that 
\begin{equation}
\small
\label{eq:tau_l+1_stiffness}
\tauM^{(l+1)^-} = K_q(\qcmd[l+1] - q^{l+1}).
\end{equation}

Combining (\ref{eq:tau_l+1_decomposition}) and (\ref{eq:tau_l+1_stiffness}) yields 
\begin{equation}
\small
\label{eq:tau_l+1_combined}
K_q(\qcmd[l+1] - q^{l+1}) = \tau_R.
\end{equation}

Finally, subtracting (\ref{eq:tau_l+1_combined}) from (\ref{eq:tau_l_combined}) gives
\begin{equation}
\small
\label{eq:stiffness_consistency_1}
K_q(q^{l+1} - q^l) = \tau_N = P_N(W) \tauM^{l^+}.
\end{equation}

As the motion from $l$ to $l+1$ needs to stay in $N(J_u)$, we need $J_u(q^{l+1} - q^l) \equiv 0$, which implies through (\ref{eq:stiffness_consistency_1}) that
\begin{equation}
\small
0 \equiv J_u K_q^{-1}P_N(W) \tauM^{l^+}, \; \forall \tauM^{l^+},
\end{equation}
which has the same form as dynamic-consistency defined in (\ref{eq:dynamic_consistency}). Not surprisingly, choosing $W = K_q$ ensures the motion generated by $\tau_N$ stays in $N(J_u)$, and the resulting pseudo-inverse $J^{K_q+}$ is called \textit{stiffness-consistent} \cite{dietrich2015overview}.

We can solve for $q^{l+1}$ by plugging (\ref{eq:tau_l_stiffness}) into (\ref{eq:stiffness_consistency_1}) and setting $W$ to $K_q$, yielding
\begin{equation}
\small
\label{eq:state_null_space_projection}
q^{l+1} = q^l + (I - J_u^{K_q+}J_u) (\qcmd[l+1] - q^l).
\end{equation}

Furthermore, as the contact force is the reaction to $\tau_R$, we have $\tau_R = -J_u^\intercal \lambda^{l+1}$, combining this with the definition of $\tau_R$ in (\ref{eq:tau_l_decomposition}) gives
\begin{equation}
\small
\label{eq:force_null_space_projection}
\lambda^{l+1} = -\left(J^{K_q+} \right)^\intercal K_q(\qcmd[l+1] - q^l).
\end{equation}

It can be shown that (\ref{eq:state_null_space_projection}) and (\ref{eq:force_null_space_projection}) are equivalent to (\ref{eq:explicit_force_and_state:state}) and (\ref{eq:explicit_force_and_state:force}), respectively. This equivalence implies that a controller based on stiffness-consistent projections and a controller (\ref{eq:generic_mpc}) using QP (\ref{eq:quasistatic_dynamics_qp}) as its dynamics constraint share the same underlying dynamics model. Therefore, the QP formulation in Sec. \ref{sec:MPC} is preferred as it can handle inequality constraints.

\section{QP Controller with Quasistatic Dynamics \label{sec:MPC}}
\subsection{Frictionless contacts}
To track a reference trajectory $q_\text{ref}(t)$ as closely as possible while respecting dynamics constraints and upper bounds on contact forces, we can specialize the generic optimization-based controller (\ref{eq:generic_mpc}) to the following QP:
\begin{subequations}
\small
\label{eq:mpc_qp1}
\begin{align}
&\underset{q^{l+1}, q_{\text{cmd}}^{l+1}, \lambda^{l+1}} {\rm min} \; \left\|q^{l+1} - q_\text{ref}^{l+1}\right\|^2 + 
\epsilon \left\|q_{\rm cmd}^{l+1} - q_\text{ref}^{l+1}\right\|^2 , \text{s.t.} \label{eq:mpc_qp1:objective} \\
&K_q(q^{l+1} - q_\text{cmd}^{l+1}) - J_u^\intercal \lambda^{l+1} = 0 \label{eq:mpc_qp:dynamics1}\\
&J_u(q^{l+1} - q^l) = 0 \label{eq:mpc_qp:dynamics2}\\
&\lambda^{l+1} \leq \lambda_\text{max} \label{eq:mpc_qp:force_upper_bound}\\
&\left|q_\text{cmd}^{l+1} - \qcmd[l]\right| \leq \Delta q_\text{max}. \label{eq:mpc_qp:velocity_upper_bound}
\end{align}
\end{subequations}
Here, the dynamics constraint (\ref{eq:generic_mpc:dynamics_constraint}) consists of (\ref{eq:mpc_qp:dynamics1}) and (\ref{eq:mpc_qp:dynamics2}), which are the KKT conditions (\ref{eq:dynamics_qp_kkt}) of the quasistatic dynamics (\ref{eq:quasistatic_dynamics_qp}). Constraint (\ref{eq:mpc_qp:force_upper_bound}) places an upper bound on contact forces. The last constraint (\ref{eq:mpc_qp:velocity_upper_bound}) bounds how quickly $\qcmd$ changes.

In the objective (\ref{eq:mpc_qp1:objective}), the first term penalizes deviation at $l+1$ from the reference trajectory. The second term, weighted by a small positive scalar $\epsilon$, adds regularization without which the objective would become semi-definite. To see why, we re-write $\left(q^{l+1} - q_\text{ref}^{l+1}\right)$ by expressing $q^{l+1}$ explicitly as a function of $q_\text{cmd}^{l+1}$ using (\ref{eq:state_null_space_projection}):
\begin{equation}
\small
\label{eq:q_minus_q_ref}
q^{l+1} - q_\text{ref}^{l+1} = (I - J_u^{K_q+}J_u) (q^{l+1}_\text{cmd} - q^l) + (q^{l} - \qref[l+1]),
\end{equation}
where the second term is a constant, and the first term multiplies $(q_\text{cmd}^{l+1} - q^l)$ by a projection which has a non-zero null space.

In the quasistatic dynamics (\ref{eq:quasistatic_dynamics_qp}), expressing contact constraints as equality constraints (\ref{eq:quasistatic_dynamics_qp:equality_constraint}) and contact forces as the constraints' Lagrange multipliers implies that the contacts are bi-lateral and frictionless. In reality, however, contacts are uni-lateral and frictional.

The bi-lateralness of (\ref{eq:quasistatic_dynamics_qp:equality_constraint}) is less concerning. As contact sensors are inevitably noisy, only contact forces above a threshold are added to QP (\ref{eq:mpc_qp1}). In addition, by (\ref{eq:explicit_force_and_state:force}), the change in contact force is bounded as long as $\left|\qcmd[l+1] - q^l\right|$ is bounded, and the boundedness of $\left|\qcmd[l+1] - q^l\right|$ is enforced by (\ref{eq:mpc_qp:velocity_upper_bound}). Therefore, as long as the bound $\Delta q_\text{max}$ is sufficiently small, we do not need to worry about contact forces flipping sign in the middle of a control step. 

On the other hand, naively ignoring friction will severely impact the performance of the controller, which motivates the mitigating measures detailed in the next subsection.

\subsection{Frictional contacts}
It is possible to model friction contact in quasistatic dynamics \cite{pang2021convex}, but control through a frictional contact requires estimating the contact normal and the friction coefficient, in addition to estimating contact forces. This requires either more sophisticated whole-arm contact sensors, or making additional assumptions about the environments that make the control-estimation pipeline more brittle. 

Therefore, we will retain the simpler frictionless contact model for controlling through a frictional contact, and mitigate the side effects of the wrong contact model by modifying the frictionless QP (\ref{eq:mpc_qp1}) to
\begin{subequations}
\small
\label{eq:mpc_qp2}
\begin{align}
&\underset{q^{l+1}, q_{\text{cmd}}^{l+1}, \lambda^{l+1}} {\rm min} \; \underbrace{\left\|q_\text{cmd}^{l+1} - q_\text{ref}^{l+1}\right\|^2}_{\text{tracking}} + \underbrace{
w^l \left\|q_\text{cmd}^{l+1} - q_\text{cmd}^{l}\right\|^2}_{\text{damping}} \label{eq:mpc_qp2:objective} \\
& \text{s.t. } \text{(\ref{eq:mpc_qp:dynamics1}), (\ref{eq:mpc_qp:dynamics2}), (\ref{eq:mpc_qp:force_upper_bound}) and (\ref{eq:mpc_qp:velocity_upper_bound}),} \nonumber
\end{align}
\end{subequations}
which has the same constraints as (\ref{eq:mpc_qp1}) but a different objective. In the rest of this section, we will elaborate on the reason for both terms in the objective (\ref{eq:mpc_qp2:objective}).

\subsubsection{Tracking}
\begin{figure}[h]
\centering
\includegraphics[width=0.80\linewidth]{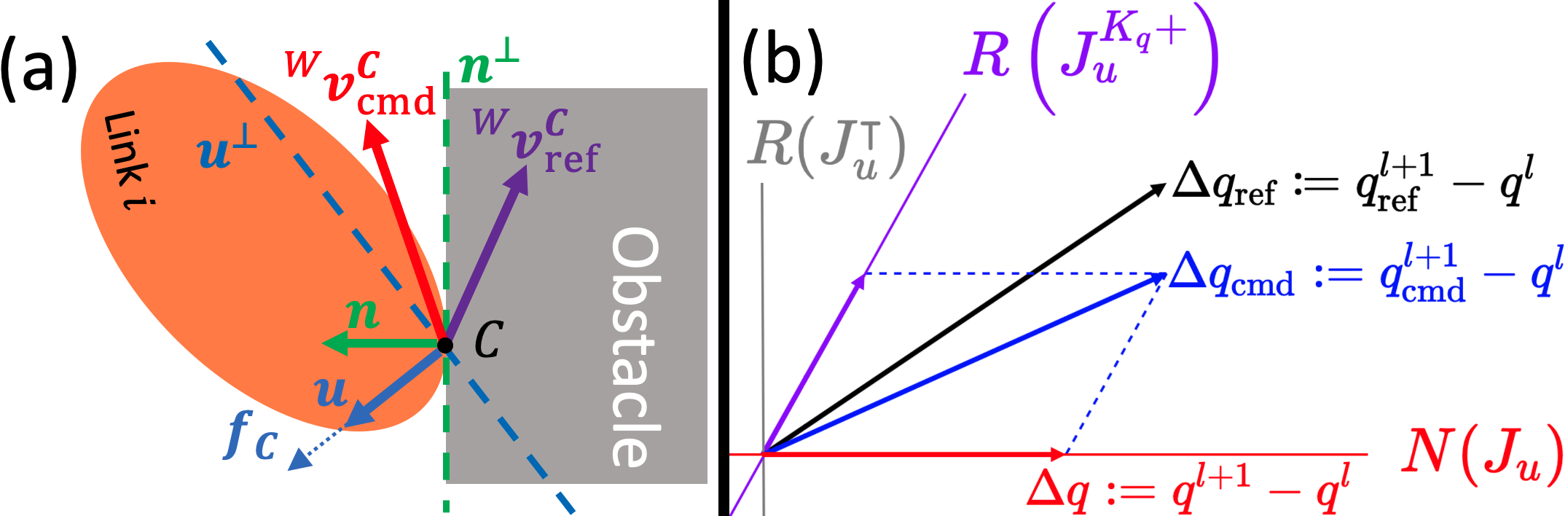}
\caption{\textbf{(a)}: The reference trajectory brings the robot into contact at $C$. Due to friction, the contact normal $n$ and the contact force direction $u$ are different. Therefore, the commanded velocity at $C$ can separate from the obstacle even when the angle between $u$ and ${}^W v^C_\text{cmd}$ is greater than $\pi / 2$. \textbf{(b)}: Definitions of $\Delta q_\text{ref}$, $\Delta q_\text{cmd}$ and $\Delta q$. The range and null space of the projection $I - J_u^{K_q+}J_u$ are $N(J_u)$ and $R\left(J_u^{K_q+}\right)$, respectively.}
\label{fig:frictional_contact_objective}
\end{figure}

When the reference trajectory $q_\text{ref}^{l+1}$ leads the robot to make contact with a frictional surface at point $C$ (Fig. \ref{fig:frictional_contact_objective}a), the surface normal $n \in \mathbb{R}^3$ and the contact force direction $u \in \mathbb{R}^3$ can be different. However, the frictionless contact model (\ref{eq:mpc_qp:dynamics1})-(\ref{eq:mpc_qp:dynamics2}) assumes that $u$ is always the same as $n$. Therefore, it is possible for ${}^W v^C_\text{cmd}$, the commanded velocity of $C$, to have a negative component along $u$ but a positive component along $n$, as shown in Fig. \ref{fig:frictional_contact_objective}a. Such a ${}^W v^C_\text{cmd}$ would lead to the robot separating from the obstacle at $l+1$. When the frictionless QP (\ref{eq:mpc_qp1}) is constructed again at $l+1$, no contact force constraints are added but $q_\text{ref}^{l+2}$ can still lead the robot to contact with a large amount of penetration. Therefore, the robot could re-establish contact with a large contact force at $l+2$.

Although it is difficult to guarantee that ${}^W v^C_\text{cmd}$ has a negative component along $n$ without knowing $n$, undesired contact jitters can be effectively reduced by making ${}^W v^C_\text{cmd}$ as close as possible to ${}^W v^C_\text{ref}$. In joint space, this translates to minimizing the distance between $\qcmd[l+1]$ and $\qref[l+1]$, which can be achieved by replacing the term $\left\|q^{l+1} - q_\text{ref}^{l+1}\right\|^2$ in (\ref{eq:mpc_qp1:objective}) by $\left\|q_\text{cmd}^{l+1} - q_\text{ref}^{l+1}\right\|^2$ in (\ref{eq:mpc_qp2:objective}).

To further illustrate the advantage of the new objective, we first re-write $q_\text{cmd}^{l+1} - q_\text{ref}^{l+1}$ using the relative quantities defined in Fig. \ref{fig:frictional_contact_objective}b:
\begin{equation}
\small
q_\text{cmd}^{l+1} - \qref[l+1] = \left(q_\text{cmd}^{l+1} - q^l \right) - \left(\qref[l+1] - q^l \right) = \Delta q_\text{cmd}^{l+1} - \Delta \qref[l+1].
\end{equation}

It is also easy to see from (\ref{eq:q_minus_q_ref}) that
\begin{equation}
\small
\Delta q^{l+1} = (I - J_u^{K_q+}J_u) \Delta q_\text{cmd}^{l+1}.
\end{equation}
The first term in the original objective (\ref{eq:mpc_qp1:objective}) thus becomes
\begin{equation}
\small
\begin{aligned}
q^{l+1} - \qref[l+1] =& \Delta q^{l+1} - \Delta \qref[l+1] \\
=& (I - J_u^{K_q+}J_u) \Delta q_\text{cmd}^{l+1} - \Delta \qref[l+1],
\end{aligned}
\end{equation}
where $(I - J_u^{K_q+}J_u)$ is the projection into $N(J_u)$ along $R\left(J_u^{K_q+}\right)$, as shown in Fig. \ref{fig:frictional_contact_objective}b. 

As a result, minimizing $\norm{\Delta q_\text{cmd}^{l+1} - \Delta \qref[l+1]}^2$ encourages $\Delta q_\text{cmd}^{l+1}$ to be close to $\Delta \qref[l+1]$ in the entire vector space. In contrast, when $\norm{\Delta q^{l+1} - \Delta \qref[l+1]}^2$ is used as the cost, only the distance between $\Delta \qref[l+1]$ and the component of $\Delta q_\text{cmd}^{l+1}$ along $N(J_u)$ is minimized.

\begin{figure*}[t!]
\vspace{0.2cm}
\centering
\includegraphics[width=1.0\textwidth]{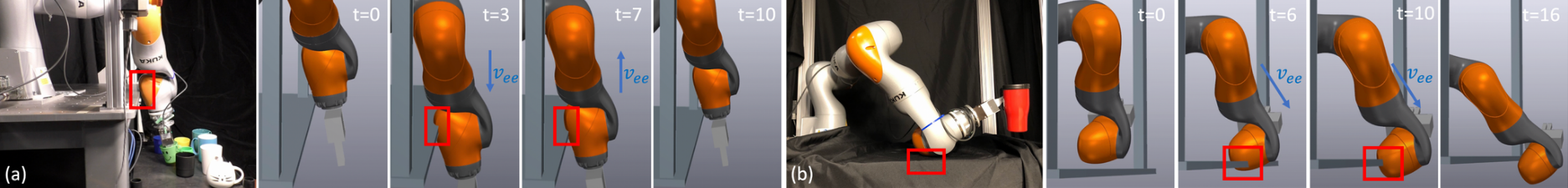}
\caption{(\textbf{a}): mug placement task. (\textbf{b}): mug moving task. Contacts are highlighted in red boxes. Blue arrows denote the direction of end effector velocity. For both tasks, the photographs on the left show where the real robot makes contact while executing the task. With collision geometry disabled, the simulation frames on the right show how much penetration would happen if the original trajectory were strictly followed. Videos of the real robot executing the tasks are included in the attachment.}
\label{fig:experiments}
\vspace{-0.6cm}
\end{figure*}

\subsubsection{Damping} 
The goal of this term is to command more conservative robot motions when we are less confident in the correctness of the frictionless contact model. As real-world contacts are almost always frictional, the contact force predicted by the frictionless model and the actual contact force measured by contact sensors are bound to be different. At every time step, this discrepancy can be quantified by
\begin{equation}
\small
e_\lambda^l \coloneqq 1 - \exp{\left(\left\|\lambda_{\rm pred}^l -\lambda_{\rm est}^l \right\|_{\infty}/a\right)} \in [0, 1],
\end{equation}
where $\lambda_{\rm pred}^l$ is the contact forces predicted by the frictional QP (\ref{eq:mpc_qp2}) at time step $l - 1$; $\lambda_{\rm est}^l$ is the measured contact forces at time step $l$; $a$ is a positive constant that weights the force prediction error. The discrepancy $e_\lambda$ is close to $1$ when the force prediction error is large, and close to $0$ when the error is small.

The weight of the second term of (\ref{eq:mpc_qp2:objective}), $w^l$, is the low-pass-filtered version of the discrepancy $e_\lambda^l$:
\begin{equation}
\small
w^l = w_\text{max} \left[ \alpha e_\lambda^l + (1 - \alpha) e_\lambda^{l-1} \right],
\end{equation}
where $w_\text{max}$ is the upper bound on $w^l$ and $\alpha$ is the forget rate of the low-pass filter. A larger $w^l$ encourages more conservative robot motions by more heavily penalizing the change in $\qcmd$ from $l$ to $l+1$.

\section{Experiments}
In this section, we demonstrate the advantages of the proposed contact-aware controller through two tasks that involve unexpected contact with the environment, which are shown in Fig. \ref{fig:experiments}. In both experiments, $\qcmd$ is tracked using the iiwa's factory impedance controller. The factory controller's stiffness is set to $[800, 600, 600, 600, 400, 200, 200]$ $\mathrm{N \cdot m/rad}$, from base joint to wrist joint. The proposed contact-aware controller runs at 200Hz. The external contacts are estimated from iiwa's external torque measurements using the Contact Particle Filter \cite{manuelli2016localizing}, which runs at around 100Hz. QPs are constructed using Drake's  \texttt{MathematicalProgram} interface \cite{drake} and solved by GUROBI \cite{gurobi}. 

In order to reduce sensitivity to measurement noise, only contact forces with norm $f_i \geq f_\text{threshold}$ are considered when constructing the contact Jacobian $J_u$ (\ref{eq:contact_jacobian}). We have chosen $f_\text{threshold}=5\mathrm{N}$, and set the upper bound on contact forces in (\ref{eq:mpc_qp:force_upper_bound}) to be $\lambda_{\text{max}}=15\mathrm{N}$. Ignoring contacts with small contact forces can be justified by the passivity of the robot's internal controller \cite{albu2007unified}, which ensures stability in the presence of external contacts.

\subsection{Mug placement task (Fig. \ref{fig:experiments}a)}
This task is defined by an end-effector pose trajectory $\left({}^W R^{T_r} (t), {}^W p^{T_r}(t)\right)$, where $T_r$ is the reference for the tool frame $T$, ${}^W R^{T_r}$ is the orientation of frame $T_r$ w.r.t. the world frame $W$, and $ {}^W p^{T_r}$ is the position of the origin of $T_r$ in world frame. It is straightforward to modify the tracking term in the frictional QP objective (\ref{eq:mpc_qp2:objective}) to minimize the pose difference between frame $T$ and its reference $T_r$, as described in \cite{koolen2016design} \cite[Chapter 3]{tedrake2021manipulation}.

The robot starts with a mug held in the gripper. It then (\textbf{i}) reaches down ($-z$ of world frame) by $0.22m$ in 4s, (\textbf{ii}) opens the gripper and drops the mug on the cart below the table in 2s, and (\textbf{iii}) moves back up to where it started in 4s. The orientation of the gripper is kept constant throughout the trajectory. As shown in Fig. \ref{fig:experiments}a, the ``wrist" (link 6) of the robot collides with the edge of the table as the gripper moves down and up.

\subsubsection{Contact force}
\begin{figure}
\centering
\includegraphics[width=0.98\linewidth]{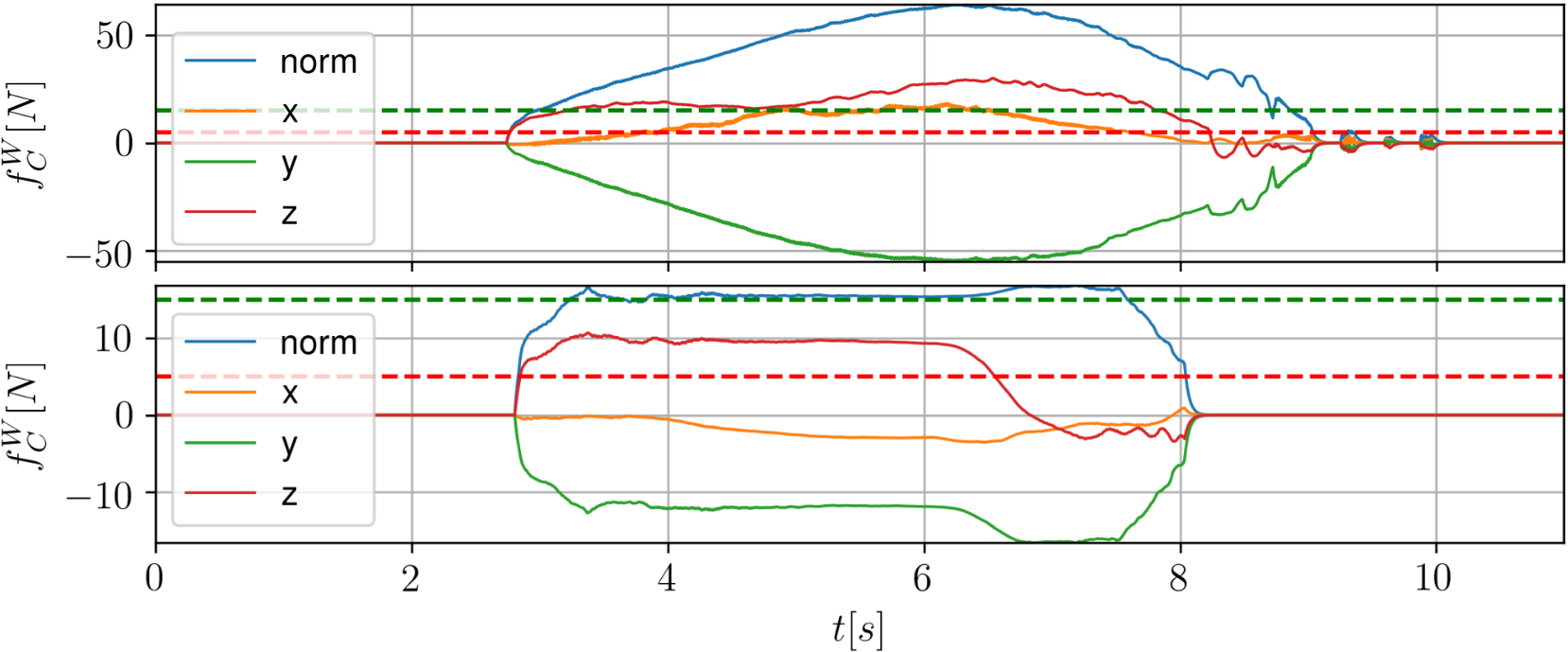}
\caption{ \textbf{mug placement task}. $x$, $y$, $z$ components and the norm of the contact force $\bm{f}_W^C$. \textbf{Top}: the baseline controller without contact force upper bounds is shown in the top plot. \textbf{Bottom}: the contact-aware controller (\ref{eq:mpc_qp2}) with a modified end effector tracking objective. In both plots, the horizontal red dashed line represents $f_\text{threshold}$ and the green dashed line $\lambda_\text{max}$.}
\label{fig:mug_placement_damped_lsq_vs_ours}
\end{figure}

As shown in Fig. \ref{fig:mug_placement_damped_lsq_vs_ours}, except during the initial impact, our controller is able to keep the contact force norm $\norm{\quantity{f}{}{C}{W}}$ close to $\lambda_\text{max}$. In contrast, the baseline controller we compare against, which computes $\qcmd$ by greedily minimizing tracking error without the dynamics and contact force constraints (\ref{eq:mpc_qp:dynamics1})-(\ref{eq:mpc_qp:velocity_upper_bound}), incurs larger contact forces.

\subsubsection{Tracking error}
\begin{figure}
\vspace{-0.2cm}
\centering
\includegraphics[width=1.0\linewidth]{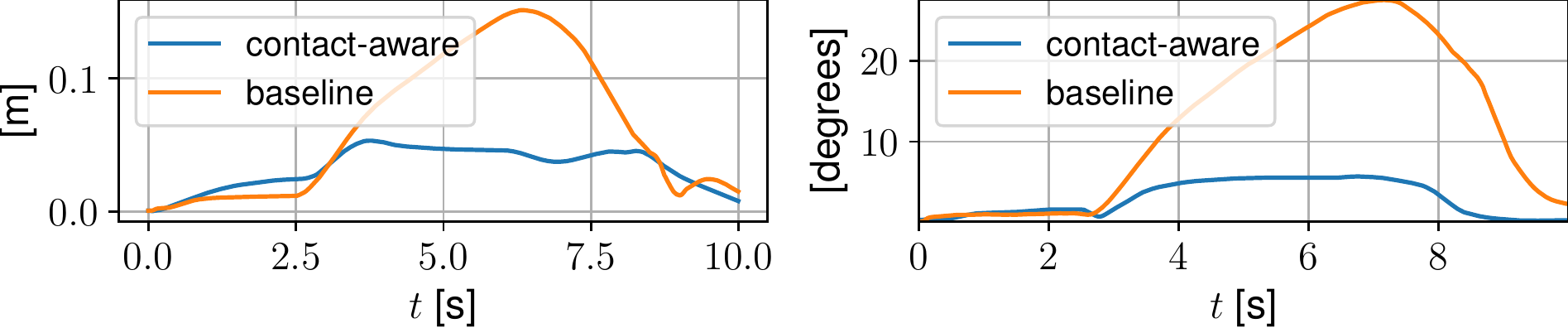}
\caption{\textbf{mug placement task}. \textbf{Left}: position tracking error. \textbf{Right}: orientation tracking error. }
\vspace{-0.6cm}
\label{fig:mug_placement_tracking_error}
\end{figure}

As shown in Fig. \ref{fig:mug_placement_tracking_error}, compared with the baseline, the contact-aware controller produces significantly less tracking error in the presence of external contact.

\vspace{-0.1cm}
\subsection{Mug moving task (Fig. \ref{fig:experiments}b) \label{sec:mug_holding_results}} 
This task is defined by a joint-space trajectory $q_{\text{ref}}(t)$, with the goal of moving the mug along a straight line while keeping the mug orientation constant. The 16s trajectory $q_{\text{ref}}(t)$ is an interpolation between joint-space knot points obtained by inverse kinematics. As shown in Fig. \ref{fig:experiments}b, to move the mug to the desired destination, the robot needs to first establish a contact with the top face of the table, and then breaks contact with the side of the table.

\begin{figure}[h]
\vspace{-0.2cm}
\centering
\includegraphics[width=0.98\linewidth]{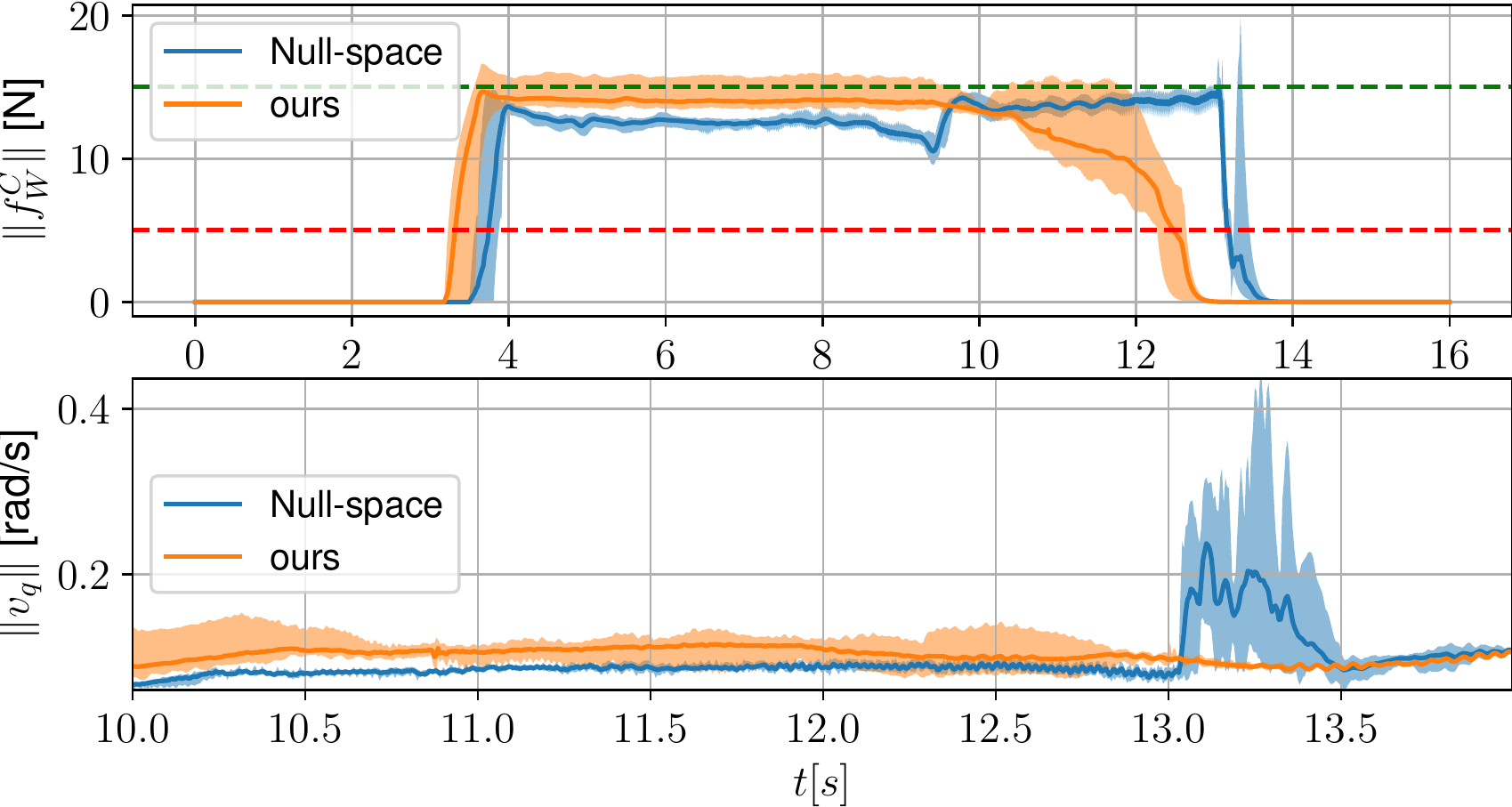}
\caption{Comparison between our QP controller and null-space projection-based control in the \textbf{mug moving task}. \textbf{Top}: contact force norm $\norm{f_W^C}$. The red dashed line represents $f_\text{threshold}$ and the green dashed line $\lambda_\text{max}$. Both controllers can keep $\norm{f_W^C}$ near $\lambda_\text{max}$ when the robot is in contact. \textbf{Bottom}: robot joint velocity norm $\norm{v_q}$ during contact separation (from $t=10s$ to $t=14s$). Note the velocity spike in the null-space projection controller. In both plots, the solid lines are the mean of 10 runs; the shaded regions around the lines represent the maximum and minimum values of all runs.}
\label{fig:jorda_vs_ours}
\vspace{-0.7cm}
\end{figure}

The baseline we are comparing against is the class of controllers trying to achieve similar goals as ours but uses null-space projection, such as \cite{jorda2019contact}. As null-space projection cannot handle inequality constraints, the contact force upper bound is usually enforced by an equality constraint which sets the contact force to $\lambda_\text{max}$. When commanded to break contact by the reference trajectory, i.e.
\begin{equation}
\small
\label{eq:break_away}
    J_u (q_\text{ref}^{l+1} - q^l) \leq 0,
\end{equation}
the contact force constraint needs to be abruptly removed in order to continue to track the reference trajectory \cite{jorda2019contact}. This can lead to large joint velocity during separation, as shown in Fig. \ref{fig:jorda_vs_ours}.

In contrast, the proposed QP controller does not explicitly make the decision to break contact based on (\ref{eq:break_away}), instead the break of contact comes naturally as a consequence of solving QP (\ref{eq:mpc_qp2}) with the inequality constraints on contact forces (\ref{eq:mpc_qp:force_upper_bound}). As shown in Fig. \ref{fig:jorda_vs_ours}, when the robot separates from the table, the drop in $\norm{f_W^C}$ occurs gradually with our QP controller, but abruptly with a null-projection-based controller.

\section{Conclusions}
We have presented a contact-aware controller that reconciles trajectory tracking with safety in unexpected contacts. The proposed controller is formulated as a QP with a quadratic cost on tracking error, a quasi-static model of the robot dynamics as constraints, and upper bounds on contact forces.

The tasks for hardware experiments are designed based on our vision of future motion planners: they are comprised of smooth, simple trajectories defined by only a few knot points. We have shown that the proposed controller is able to keep both the tracking error and contact forces small if the robot makes an accidental contact. In addition, our controller outperforms controllers based on null-space projection when an established contact needs to be broken as the robot follows a reference trajectory.

It is difficult to reliably sense more than one contacts on the arm from only joint torque \cite{pang2021identifying}. Nevertheless, with a more capable contact sensor such as \cite{luo2021learning}, we believe the proposed contact-aware controller will greatly reduce robots' reliance on environment sensing/monitoring and collision-free motion planning.

\clearpage
\bibliographystyle{IEEEtran}
\bibliography{references.bib}

\end{document}